\documentclass[conference]{IEEEtran}
\IEEEoverridecommandlockouts
\usepackage{cite}
\usepackage{amsmath,amssymb,amsfonts}
\usepackage{algorithmic}
\usepackage{graphicx}
\usepackage{textcomp}
\usepackage{xcolor}
\usepackage{multirow}
\usepackage{booktabs}
\usepackage{amssymb} 
\usepackage{graphicx} 
\usepackage{multirow}
\usepackage{booktabs}
\usepackage[colorlinks=true, allcolors=blue]{hyperref}
\usepackage{tabularx} 

\begin{document}

\title{Deepfake Detection: Leveraging the Power of 2D and 3D CNN Ensembles}
\author{
    Aagam Bakliwal\textsuperscript{1}, 
    Amit D. Joshi\textsuperscript{2} \\
    \textsuperscript{1,2}Department of CSE, COEP Technological University (COEP Tech) \\
    Pune, Maharashtra, India
}
\date{}

\maketitle

\begin{abstract}
In the dynamic realm of deepfake detection, this work presents an innovative approach to validate video content. The methodology blends advanced 2-dimensional and 3-dimensional Convolutional Neural Networks. The 3D model is uniquely tailored to capture spatiotemporal features via sliding filters, extending through both spatial and temporal dimensions. This configuration enables nuanced pattern recognition in pixel arrangement and temporal evolution across frames. Simultaneously, the 2D model leverages EfficientNet architecture, harnessing auto-scaling in Convolutional Neural Networks. Notably, this ensemble integrates Voting Ensembles and Adaptive Weighted Ensembling. Strategic prioritization of the 3-dimensional model's output capitalizes on its exceptional spatio-temporal feature extraction. Experimental validation underscores the effectiveness of this strategy, showcasing its potential in countering deepfake generation's deceptive practices.
\end{abstract}


\section{Introduction}
\begin{figure}
  \centering
  \includegraphics[width=\columnwidth]{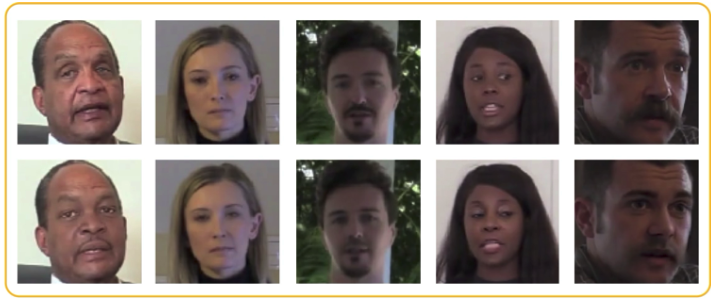}
  \caption{Paired fake samples generated from original pristine faces in the DFDC dataset}
  \label{fig:DFDC}
\end{figure}

The accelerating pace of digital innovation has propelled us into an age where content generation and distribution are more powerful than ever. While these advancements have enriched the creative landscape and expanded avenues for communication, they have also given rise to issues that jeopardize the integrity of factual information. Chief among these concerns is the advent of deepfake videos—hyper-realistic, such as in Fig.  \ref{fig:DFDC}, yet entirely fabricated visual narratives capable of causing misinformation, character assassination, and a host of other societal perils \cite{impact1}\cite{impact2}.

With the increasing sophistication of deepfakes, the demand for robust detection techniques has never been higher. Convolutional Neural Networks (CNNs), particularly 2D variants, have long been the cornerstone of image recognition tasks but are increasingly viewed as insufficient for capturing the temporal nuances inherent in video data. This realization has catalyzed the development of 3D CNNs, specifically designed to integrate both spatial and temporal dimensions, making them ideal for video-based analysis \cite{Roy2022}.

While various methods such as fingerprinting, chrominance properties have shown success in identifying GAN-based deepfakes \cite{6}\cite{7}\cite{5}, this research aspires to build a universally applicable detection model. To this end, we introduce a cutting-edge ensemble model that synergizes the strengths of both 2D and 3D CNNs. The proposed approach melds the efficient spatial feature extraction capabilities of 2D CNNs—particularly focusing on the EfficientNet architecture—with the spatiotemporal capabilities of 3D CNNs. By doing so, this research aims for a more comprehensive analysis of video data that can unearth subtle patterns from both individual frames and their temporal relationships. In addition, this model incorporates advanced ensemble techniques such as Voting Ensembles \cite{ensemble1} and Adaptive Weighted Ensembling \cite{ensemble2} to optimize detection accuracy.

This work serves as a comprehensive guide for those looking to navigate the multifaceted challenges of deepfake detection which aims to strengthen the defenses against digital falsehoods and ensure that the digital landscape remains a bastion of veracity.

The remainder of this paper is organized as follows: Section II provides an overview of related work in the domain. The methodology adopted for deepfake detection is elaborated in Section III. Section IV describes the experiments conducted, followed by a detailed discussion of results in Section V. The paper concludes with insights and future directions in Section VI.

\section{Related Work}
The field of deepfake detection has been enriched with numerous techniques, both from deep learning and traditional realms. While various non-deep learning approaches have been widely explored in past studies, Generative Adversarial Networks (GANs) have particularly dominated this arena. They are employed to analyze the artifacts produced by deepfakes, aiming to detect these minute inconsistencies. Studies have revealed unique chrominance properties of deepfake images distinct from camera-produced images \cite{6}\cite{7}. Their analysis, especially in the chrominance components and the residual domain, can differentiate between GAN produced imagery and authentic camera images. Furthermore, peculiarities in deepfake outputs were studied, and different GAN sources were identified \cite{5}. This underscores the significance of anomalies in images produced by GANs for deepfake detection. However, these methods primarily target GAN-generated deepfakes, overlooking other types of deepfakes, thereby compromising their robustness.

Support Vector Machines (SVMs) have also found extensive application in deepfake detection. For instance, employed facial geometry alongside SVM classification have been used to detect anomalies like unnatural reflections and detailed discrepancies in eyes and teeth. Using distinct feature vector sets, they exposed deepfakes' challenges in achieving lifelike outputs \cite{8} . Other studies utilized SVM-based classification, the former leveraging a pixel anomaly-based feature extraction technique and the latter employing the Speeded Up Robust Features (SURF) algorithm with the Bag of Words (BoW) model for detecting face swaps \cite{3}\cite{4}.

Frequency-based techniques have been another promising direction. A compression-oriented approach using a self-supervised decoupling network \cite{16} and harnessing harnessed a ResNet-based Pixel CNN \cite{15} have been proposed. However, their performance fluctuated with different compression rates. The performance fluctuation observed with different compression rates indicates a potential challenge in maintaining consistent detection accuracy across various data settings. The potential of the phase spectrum, particularly its sensitivity to upsampling, was explored using the Discrete Fourier Transform (DCT) \cite{18}. Several studies have also explored the utility of frequency-aware frameworks and high-frequency noise features for the detection of deepfakes \cite{19}\cite{20}. While these methods have demonstrated efficacy in well-lit scenarios, they exhibit sensitivity to environmental noise that adversely impact the high-frequency components crucial for detection.

Various innovative deep learning techniques have also been proposed for deepfake detection. A mouth-centric approach was the focus of one study, although it encountered difficulties with videos where the mouth was obscured \cite{13}. Another research effort highlighted the employment of biological signals \cite{12}. Principal component analysis (PCA) and Convolutional LSTM Residual Network (CLRNet) were explored in other studies \cite{17}\cite{14}. Yet another study combined eye sequences with a hybrid network that integrates VGG and LSTM architectures, although it faced challenges related to the natural variability in human eye blinking \cite{11}.

Additionally, methods centered around neural networks have also been developed. FakeSpotter was introduced in one such study; it used a shallow-layered architecture to scrutinize neuron behavior and outperformed other detectors in terms of accuracy \cite{21}. However, its shallow architecture may limit its ability to capture more complex features of deepfakes.  Other research made use of pre-trained networks, with one specifically focusing on the XceptionNet and distortions in facial regions in deepfakes \cite{22}\cite{23}. Inflated 3D ConvNets (I3D) \cite{i3d} have also been used, which capture important spatio-temporal features of videos, which can be used to identify deepfakes. 

Building upon the methodologies presented in \cite{thispaper} and \cite{i3d}, the aim of this work is to formulate a more robust architecture for deepfake detection. The model is engineered for resiliency, capable of identifying a wide array of deepfakes beyond those generated through GANs or other deep learning techniques. By integrating a 3D CNN into our framework, we strive to enhance the overall performance of deepfake detection.

\section{Proposed Method}

This section elucidates our approach for deepfake detection, rooted in the principle of ensembling. The power of model ensembling, renowned for its potential to amplify prediction performance, is harnessed in this method. Our central aim is to explore the feasibility and methodologies of training diverse CNN-based classifiers—encompassing both 2D and 3D - to extract complementary high-level semantic information. Such harmonized information sources are anticipated to synergistically enhance the ensemble's performance for deepfake detection. Concurrently, we place an emphasis on devising a model that remains both nimble in design and straightforward to train.

\subsection{Attention2D}

This method's 2D model capitalizes on the strengths of the EfficientNet model series, a groundbreaking approach for CNN automatic scaling, known for its high accuracy and efficiency \cite{19}. We specifically selected the EfficientNetB4 architecture due to its optimal balance between model parameters, computational cost, and classification prowess, as highlighted in \cite{19}. In comparison to XceptionNet, a face manipulation detection standard presented in \cite{xception}, EfficientNetB4 achieved higher top-1 accuracy with 4 million fewer parameters and a reduction of 4.2 Billion FLOPS. This led to the preference for the EfficientNetB4.

\begin{figure*}  
  \centering
  \includegraphics[width=\linewidth]{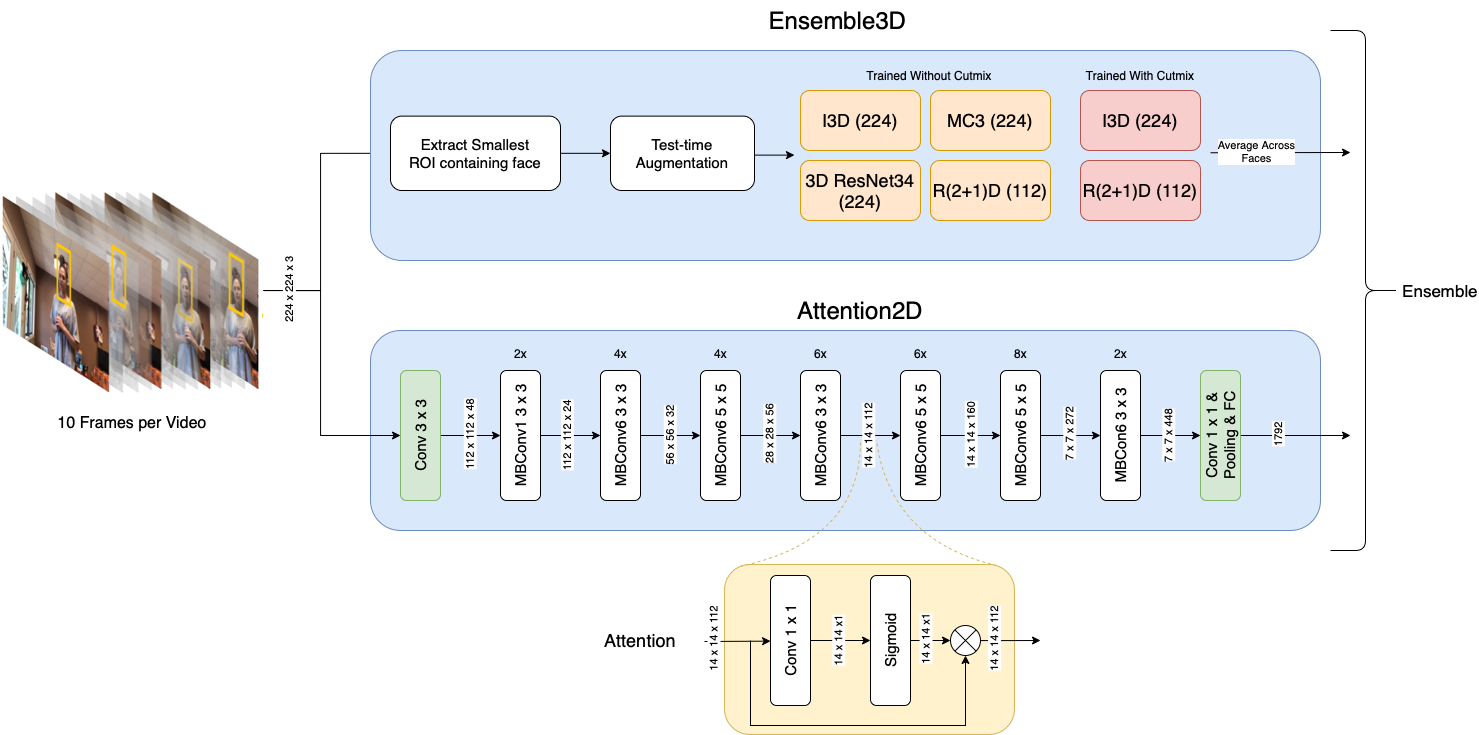}
  \caption{Proposed Model Workflow}
  \label{fig:efficient}
\end{figure*}

The EfficientNetB4 structure is outlined within the blue block of Fig. \ref{fig:efficient}, consistent with the definitions provided in \cite{19}. The input to the network is squared facial images taken from individual video frames, efficiently extracted through robust face detectors mentioned in \cite{31}\cite{32}.

Expanding upon this foundation, this method implements modifications drawing inspiration from both the realms of natural language processing and computer vision by incorporating attention mechanisms. Influential works such as the residual attention networks \cite{33} and transformers \cite{20} have showcased the ability of neural networks to focus on the most pertinent parts of their input, be it imagery or textual sequences. This integrated attention approach merges the inherent attention mechanism of EfficientNet with self-attention techniques from previous studies \cite{29}\cite{34}. This procedure is as follows:
\begin{enumerate}
\item Extract feature maps, sized 14 × 14 × 112, from EfficientNetB4 up to its fourth MBConv block.
\item Process these feature maps with a singular convolutional layer of kernel size 1, subsequently applying a Sigmoid activation function to yield a single attention map, as advised in \cite{thispaper}.
\item Multiply the resulting attention map element-wise with the feature maps from the designated layer.
\end{enumerate}

For visualization, this attention methodology is depicted in the yellow block of Fig. \ref{fig:efficient}.

This attention mechanism has a two-fold advantage. It not only guides the network to emphasize significant areas of the feature maps but also reveals which parts of the input it deems most valuable. The ensuing attention map can overlay the input, marking areas deemed critical by the network. The attention block's output is then directed to the rest of the EfficientNetB4 layers. The augmented network is termed as Attention2D and undergoes end-to-end training.

\subsection{Ensemble3D}

For each face within the video, this method specifically focus on the smallest region of interest, ensuring that our model emphasizes facial features that could be indicative of deepfake manipulations. This methodology is particularly essential given that deepfake techniques often exhibit subtle temporal inconsistencies, which might be missed in standard 2D convolution approaches.

The 3D convolution technique stands out from its 2D counterpart primarily because of its capacity to slide across an additional dimension—time (or depth) alongside height and width. While 2D convolutions operate on individual frames, 3D convolutions understand videos as volumetric data, adding the crucial temporal dimension. Consequently, the filters in a 3D convolution evolve to be three-dimensional, thereby allowing the model to recognize patterns that transcend individual frames, capturing anomalies over a sequence of frames, and offering a more holistic view of the video \cite{3DAdv}.

This method leverages four distinct 3D convolution models. The first is the I3D model \cite{i3d}, an innovation that ingeniously employs sets of RGB frames as input. This model is a modification of the acclaimed Inception model, with its 2D convolutional layers transformed into 3D to tap into the potential of spatio-temporal modeling. To kickstart training, the pre-trained weights of the Inception model trained on ImageNet are inflated. Prior research has substantiated that such inflation proposed approach leads to improved 3D images generation quality, providing a significant boost to the performance of 3D models \cite{InflationBenefit}.

Supplementing the I3D model, three other models are deployed: 3D ResNet34, MC3 and R(2+1)D \cite{resnets}. To fortify the models and ensure they don't develop a dependency on specific facial features or regions, the CutMix data augmentation technique is employed  with I3D and R(2+1)D. CutMix's prowess lies in its ability to compel models to focus on diverse regions of the input, thereby fostering a more robust recognition system \cite{CutMixPaper}.

Finally, rather than depending on a singular model, this method amalgamates the strengths of all trained models into an ensemble. This approach capitalizes on the unique strengths and mitigates the individual weaknesses of each model. The predictions of this ensemble are then averaged across all faces within a video, ensuring a comprehensive and reliable evaluation.

\subsection{Ensembling Techniques}
In this approach, we advocate for a synergistic fusion of Voting Ensembles \cite{ensemble1} and Adaptive Weighted Ensembling \cite{ensemble2}. This blend aims to harness the strengths of individual ensemble techniques for a robust prediction performance. Within this ensemble framework, the results derived from the 3D model are assigned a higher weight, recognizing its inherent ability to discern spatio-temporal features, which are paramount in video-based deepfake detection. Delving deeper into the internal workings of both our 3D and 2D models, Voting Ensembles are employed as the cornerstone mechanism to aggregate and finalize predictions. This ensemble within an ensemble strategy not only fortifies the prediction robustness but also ensures that the intricacies of the data are well-captured and represented in the final outcomes.

The proposed workflow is shown in Fig. \ref{fig:efficient}.

\section{Experiments}

This sections reports all the details regarding the datasets used and the experimental setup.

\subsection{Dataset}
The method under consideration is evaluated using the DFDC dataset, which was specifically released for a corresponding Kaggle competition. Comprising over 119,000 video sequences designed for this challenge, the dataset includes both genuine and manipulated videos. Authentic videos feature a diverse range of actors in terms of gender, skin tone, age, and so on, and are recorded against a variety of backgrounds to add visual complexity. On the other hand, the manipulated videos are generated from these real videos through various DeepFake techniques, such as multiple face-swapping algorithms. Each sequence has an approximate length of 300 frames. Notably, the dataset is heavily skewed towards fake videos, containing around 100,000 fake sequences and 19,000 real ones.

\subsection{Network Comparison}
The experimental setup considers a variety of neural networks for benchmarking:

\begin{enumerate}
\item As the top-performing model cited in \cite{17}, XceptionNet serves as an essential point of reference for our tests.
\item EfficientNetB4 stands out for its superior accuracy and efficiency relative to other methodologies, as documented in \cite{19}.
\item EfficientNetB4Att excels as the highest-performing model in the scope of this research, as mentioned in \cite{thispaper}.
\item A streamlined 3D convolutional neural network specifically designed for deepfake identification is introduced in \cite{comparePaper1}.
\end{enumerate}

For each network, individual training and evaluation are conducted using the DFDC dataset.

\subsection{Setup}
From each video, 10 frames are selected. This choice is grounded in the understanding that increasing the frame count per video can lead to overfitting, and any additional frames don't significantly enhance performance as per \cite{thispaper}.

The analysis conducted by primarily centers on areas featuring the subject's face. As a preliminary step, BlazeFace extractor \cite{32} is employed to isolate faces from each frame, which proves to be speedier in their trials compared to the MTCNN detector \cite{31} utilized in \cite{17}. In cases where multiple faces are detected, the face with the highest confidence rating is retained. The subsequent input to the neural network is a squared color image with dimensions of 224 × 224 pixels. 

To bolster the resilience of the models during training and validation, data augmentation techniques are implemented on the extracted faces. Specifically, random downscaling, horizontal flipping, adjustments in brightness and contrast, hue saturation, shearing, rotations, noise addition, and normalization are employed by leveraging Albumentation \cite{36} for these tasks.

For the training phase, the Adam optimizer \cite{38} is chosen with tailored hyperparameters: $\beta_1 = 0.9$, $\beta_2 = 0.999$, $\epsilon = 10^{-8}$, with an initial learning rate at $10^{-5}$. This combination of optimizer and hyperparameters plays a pivotal role in the successful training of the models.

\subsection{Metrics}

\subsubsection{Log Loss}
Given an input sample face, the network produces a score, $\hat{y}_i$, correlated with the face. It's important to note that this score hasn't been subjected to a Sigmoid activation function. The process of weight updating employs the widely-accepted LogLoss function \cite{logloss}:
\begin{equation}
    \label{loglossEq}
    \mathcal{L}_{\text{log}} = -\frac{1}{N} \sum_{i=1}^{N} \left[ y_i \log (S(\hat{y}_i)) + (1-y_i) \log (1-S(\hat{y}_i)) \right]
\end{equation}

In the expression labeled as \ref{loglossEq}, $\hat{y}_i$ signifies the score attributed to the face indexed by $i$, and $y_i \in {0, 1}$ indicates the label assigned to that specific face. To elaborate, a label of 0 pertains to faces derived from authentic videos, while a label of 1 is linked to faces sourced from manipulated videos. $N$ corresponds to the total count of faces utilized in the training process, and $S(\cdot)$ represents the Sigmoid function.

\subsubsection{Area Under the Curve (AUC)}
AUC is another vital indicator for assessing how well our deepfake detection models function. In contrast to LogLoss, which calculates the degree of divergence between predicted values and actual labels, the AUC quantifies the model's competence in distinguishing real faces from manipulated ones. With a scale that varies between 0 and 1, an AUC score of 0.5 indicates that the model's performance is equivalent to random chance, whereas a score of 1 signifies flawless categorization.

Let $T = \{t_1, t_2, \ldots, t_N\}$ represent the true labels and $P = \{p_1, p_2, \ldots, p_N\}$ be the predicted probabilities after applying the Sigmoid function $S(\hat{y}_i)$ to the raw scores $\hat{y}_i$. The AUC \cite{auc} can be computed as follows:
\begin{equation}
    \label{aucEq}
    \text{AUC} = \frac{1}{N_{0} \times N_{1}} \sum_{i=1}^{N} \sum_{j=1}^{N} \mathbb{I}(t_i < t_j) \mathbb{I}(p_i > p_j)
\end{equation}

In the equation labeled as \ref{aucEq}, $N_0$ and $N_1$ denote the count of authentic and altered facial instances, respectively. The symbol $\mathbb{I}(x)$ represents the indicator function, yielding a value of 1 when condition $x$ is satisfied and 0 otherwise. This metric holds an advantage in its resistance to alterations in thresholds and its ability to offer a comprehensive assessment of the model's effectiveness across various classification thresholds.

\section{Results and Discussion}

This section presents a comprehensive overview of the gathered results attained throughout the course of experimentation.

\subsection{Attention2D}

Attention2D created an attention map, similar to \cite{thispaper}. The output from the Sigmoid layer within the attention block in extracted, producing a 2D map of size 14 x 14, which we then upscale to 224 x 224 dimensions. The proposed attention mechanism adeptly accentuates intricate facial features like eyes, mouth, nose, and ears. Conversely, regions with subtle gradients which don't offer much valuable information to the network are attenuated. As multiple studies, including \cite{16}, indicate that deepfake generation often leaves artifacts around facial features, our model selectively concentrates on these pivotal areas. This selective focus curtails redundant computations, enhancing the model's efficiency.

\subsection{Detection Capability}

This section encapsulates the findings from the foundational XceptionNet network juxtaposed with the proposed model, drawing comparisons with other studies in the field as well.

\begin{table}[!htb]
    \centering
    \caption{Comparison of Model Ensembles and their Performance Metrics}
    \label{resultsTable}
    \begin{tabular}{c|c|c|c|c|c|c|c}
        \toprule
        \multirow{2}{*}{XN} & \multirow{2}{*}{E4} & \multirow{2}{*}{E4A} & \multirow{2}{*}{E4AS} & \multirow{2}{*}{3D} & \multirow{2}{*}{2D} & \multicolumn{2}{c}{Metrics} \\
        \cline{7-8}
         & & & & & & AUC & LogLoss \\
        \midrule
        $\checkmark$ & & & & & & 0.8784 & 0.4897 \\
        & $\checkmark$ & & & & & 0.8766 & 0.4819 \\
        & & $\checkmark$ & & & & 0.8642 & 0.5133 \\
        & $\checkmark$ & $\checkmark$ & & & & 0.8785 & 0.4731 \\
        & & & $\checkmark$ & & & 0.836 & 0.5507 \\
        & $\checkmark$ & $\checkmark$ & $\checkmark$ & & & 0.8751 & 0.4717 \\
        & & & & $\checkmark$ & & 0.8821 & 0.4741 \\
        & & & & & $\checkmark$ & 0.8796 & 0.4691 \\
        & & & & $\checkmark$ & $\checkmark$ & \textbf{0.8969} & \textbf{0.4641} \\
        \bottomrule
    \end{tabular}
    \vspace{0.5em} \\
\end{table}

In Table \ref{resultsTable}, the performance metrics across various model architectures are delineated. For clarification: XN symbolizes XceptionNet, E4 represents EfficientNetB4, E4A is for EfficientNetB4Att, E4AS denotes EfficientNetB4AttST  \cite{thispaper}. 3D signifies the Ensemble3D, and 2D stands for Attention2D.

A discerning look at these outcomes reveals the inherent benefits of model ensembling with regard to performance metrics. As anticipated, optimal results often emanate from integrating two or more networks, implying that such combinations bolster both deepfake detection accuracy, which is measured via AUC, and detection quality, which is assessed through LogLoss. Notably, for the dataset in consideration, both LogLoss and AUC metrics consistently surpass the baseline values. The proposed method's Ensemble3D and Attention2D models marginally outpace other individual models, and some of their ensembles as well. Crucially, the fusion of the proposed methods 2D and 3D models eclipses prior models, testifying to its robust nature.

However, every model has its limitations. This model may require careful integration of both 2D and 3D components, making the training process potentially more complex than single-model systems. Moreover, while the ensembling approach reduces overfitting risks and captures diverse patterns, it also demands more computational resources and may be challenging to deploy in real-time scenarios or on devices with limited processing capabilities. The model's dependency on voting ensembles and weighted ensembling might also make it susceptible to potential inefficiencies if one component model underperforms.

\section{Conclusion and Future Scope}

This work puts forth an advanced system for deepfake detection, that synergistically combines the capabilities of 2D and 3D CNNs. Through the proposed ensembling strategy, focus is on harnessing the robust spatiotemporal features that 3D models are proficient at capturing, giving them higher weight in the ensemble. This focus amplifies the system's overall performance. Additionally, attention mechanisms are integrated into the 2D EfficientNet architecture, contributing not just to enhanced predictive power but also to greater model interpretability. The empirical evaluation reveals that the proposed hybrid approach significantly outpaces existing benchmarks. It achieves this by harmonizing disparate models and leveraging their complementary functionalities. It is noticeable that the performance experiences a significant enhancement when placing greater emphasis on 3D models, owing to their proficiency in handling intricate spatiotemporal relationships.

As the landscape of deepfake technology continues to grow more complex, it becomes imperative that existing detection methodologies adapt and evolve concurrently. Looking ahead, future studies could focus on incorporating even more advanced attention mechanisms, exploring a wider variety of models for ensembling, or venturing into the realm of real-time deepfake detection applications. In summary, this work offers a balanced, efficient, and interpretable avenue for tackling the evolving threat posed by deepfakes.

\bibliography{references.bib}

\begin{thebibliography}{10}
\providecommand{\url}[1]{#1}
\csname url@samestyle\endcsname
\providecommand{\newblock}{\relax}
\providecommand{\bibinfo}[2]{#2}
\providecommand{\BIBentrySTDinterwordspacing}{\spaceskip=0pt\relax}
\providecommand{\BIBentryALTinterwordstretchfactor}{4}
\providecommand{\BIBentryALTinterwordspacing}{\spaceskip=\fontdimen2\font plus
\BIBentryALTinterwordstretchfactor\fontdimen3\font minus
  \fontdimen4\font\relax}
\providecommand{\BIBforeignlanguage}[2]{{%
\expandafter\ifx\csname l@#1\endcsname\relax
\typeout{** WARNING: IEEEtran.bst: No hyphenation pattern has been}%
\typeout{** loaded for the language `#1'. Using the pattern for}%
\typeout{** the default language instead.}%
\else
\language=\csname l@#1\endcsname
\fi
#2}}
\providecommand{\BIBdecl}{\relax}
\BIBdecl

\bibitem{impact1}
\BIBentryALTinterwordspacing
J.~Currie. (2021) How deepfakes are impacting society. [Online]. Available:
  \url{https://www.ed.ac.uk/impact/opinion/how-deepfakes-are-impacting-society}
\BIBentrySTDinterwordspacing

\bibitem{impact2}
\BIBentryALTinterwordspacing
N.~University. (2022) Deepfakes and fake news pose a growing threat to
  democracy. [Online]. Available:
  \url{https://news.northeastern.edu/2022/04/01/deepfakes-fake-news-threat-democracy/}
\BIBentrySTDinterwordspacing

\bibitem{Roy2022}
R.~Roy, I.~Joshi, A.~Das, and A.~Dantcheva, \emph{3D CNN Architectures and
  Attention Mechanisms for Deepfake Detection}.\hskip 1em plus 0.5em minus
  0.4em\relax Cham: Springer International Publishing, 2022, pp. 213--234.

\bibitem{6}
\BIBentryALTinterwordspacing
H.~Li, B.~Li, S.~Tan, and J.~Huang, ``Identification of deep network generated
  images using disparities in color components,'' \emph{Signal Processing},
  vol. 174, p. 107616, 2020. [Online]. Available:
  \url{https://www.sciencedirect.com/science/article/pii/S0165168420301596}
\BIBentrySTDinterwordspacing

\bibitem{7}
F.~Matern, C.~Riess, and M.~Stamminger, ``Exploiting visual artifacts to expose
  deepfakes and face manipulations,'' \emph{2019 IEEE Winter Applications of
  Computer Vision Workshops (WACVW)}, pp. 83--92, 2019.

\bibitem{5}
\BIBentryALTinterwordspacing
F.~Marra, D.~Gragnaniello, L.~Verdoliva, and G.~Poggi, ``Do gans leave
  artificial fingerprints?'' \emph{CoRR}, vol. abs/1812.11842, 2018. [Online].
  Available: \url{http://arxiv.org/abs/1812.11842}
\BIBentrySTDinterwordspacing

\bibitem{ensemble1}
\BIBentryALTinterwordspacing
I.~E. Livieris, A.~Kanavos, V.~Tampakas, and P.~Pintelas, ``A weighted voting
  ensemble self-labeled algorithm for the detection of lung abnormalities from
  x-rays,'' \emph{Algorithms}, vol.~12, no.~3, 2019. [Online]. Available:
  \url{https://www.mdpi.com/1999-4893/12/3/64}
\BIBentrySTDinterwordspacing

\bibitem{ensemble2}
Z.~Yuan and P.~Zhao, ``An improved ensemble learning for imbalanced data
  classification,'' in \emph{2019 IEEE 8th Joint International Information
  Technology and Artificial Intelligence Conference (ITAIC)}, 2019, pp.
  408--411.

\bibitem{8}
F.~Matern, C.~Riess, and M.~Stamminger, ``Exploiting visual artifacts to expose
  deepfakes and face manipulations,'' \emph{2019 IEEE Winter Applications of
  Computer Vision Workshops (WACVW)}, pp. 83--92, 2019.

\bibitem{3}
A.~Agarwal, R.~Singh, M.~Vatsa, and A.~Noore, ``Swapped! digital face
  presentation attack detection via weighted local magnitude pattern,'' in
  \emph{2017 IEEE International Joint Conference on Biometrics (IJCB)}, 2017,
  pp. 659--665.

\bibitem{4}
Y.~Zhang, L.~Zheng, and V.~L.~L. Thing, ``Automated face swapping and its
  detection,'' \emph{2017 IEEE 2nd International Conference on Signal and Image
  Processing (ICSIP)}, pp. 15--19, 2017.

\bibitem{16}
J.~Zhang, J.~Ni, and H.~Xie, ``Deepfake videos detection using self-supervised
  decoupling network,'' in \emph{2021 IEEE International Conference on
  Multimedia and Expo (ICME)}, 2021, pp. 1--6.

\bibitem{15}
A.~Khodabakhsh and C.~Busch, ``A generalizable deepfake detector based on
  neural conditional distribution modelling,'' 09 2020.

\bibitem{18}
H.~Liu, X.~Li, W.~Zhou, Y.~Chen, Y.~He, H.~Xue, W.~Zhang, and N.~Yu,
  ``Spatial-phase shallow learning: Rethinking face forgery detection in
  frequency domain,'' in \emph{2021 IEEE/CVF Conference on Computer Vision and
  Pattern Recognition (CVPR)}, 2021, pp. 772--781.

\bibitem{19}
\BIBentryALTinterwordspacing
M.~Tan and Q.~Le, ``{E}fficient{N}et: Rethinking model scaling for
  convolutional neural networks,'' in \emph{Proceedings of the 36th
  International Conference on Machine Learning}, ser. Proceedings of Machine
  Learning Research, K.~Chaudhuri and R.~Salakhutdinov, Eds., vol.~97.\hskip
  1em plus 0.5em minus 0.4em\relax PMLR, 09--15 Jun 2019, pp. 6105--6114.
  [Online]. Available: \url{https://proceedings.mlr.press/v97/tan19a.html}
\BIBentrySTDinterwordspacing

\bibitem{20}
J.~Fridrich and J.~Kodovsky, ``Rich models for steganalysis of digital
  images,'' \emph{IEEE Transactions on Information Forensics and Security},
  vol.~7, no.~3, pp. 868--882, 2012.

\bibitem{13}
A.~Haliassos, K.~Vougioukas, S.~Petridis, and M.~Pantic, ``Lips don't lie: A
  generalisable and robust approach to face forgery detection,'' in
  \emph{Proceedings of the IEEE/CVF Conference on Computer Vision and Pattern
  Recognition (CVPR)}, June 2021, pp. 5039--5049.

\bibitem{12}
U.~A. Ciftci, I.~Demir, and L.~Yin, ``Fakecatcher: Detection of synthetic
  portrait videos using biological signals,'' \emph{IEEE Transactions on
  Pattern Analysis and Machine Intelligence}, pp. 1--1, 2020.

\bibitem{17}
\BIBentryALTinterwordspacing
A.~R{\"{o}}ssler, D.~Cozzolino, L.~Verdoliva, C.~Riess, J.~Thies, and
  M.~Nie{\ss}ner, ``Faceforensics++: Learning to detect manipulated facial
  images,'' \emph{CoRR}, vol. abs/1901.08971, 2019. [Online]. Available:
  \url{http://arxiv.org/abs/1901.08971}
\BIBentrySTDinterwordspacing

\bibitem{14}
\BIBentryALTinterwordspacing
S.~Tariq, S.~Lee, and S.~S. Woo, ``A convolutional {LSTM} based residual
  network for deepfake video detection,'' \emph{CoRR}, vol. abs/2009.07480,
  2020. [Online]. Available: \url{https://arxiv.org/abs/2009.07480}
\BIBentrySTDinterwordspacing

\bibitem{11}
Y.~Li, M.-C. Chang, and S.~Lyu, ``In ictu oculi: Exposing ai created fake
  videos by detecting eye blinking,'' 12 2018, pp. 1--7.

\bibitem{21}
\BIBentryALTinterwordspacing
R.~Wang, F.~Juefei-Xu, L.~Ma, X.~Xie, Y.~Huang, J.~Wang, and Y.~Liu,
  ``Fakespotter: A simple yet robust baseline for spotting ai-synthesized fake
  faces,'' in \emph{Proceedings of the Twenty-Ninth International Joint
  Conference on Artificial Intelligence, {IJCAI-20}}, C.~Bessiere, Ed.\hskip
  1em plus 0.5em minus 0.4em\relax International Joint Conferences on
  Artificial Intelligence Organization, 7 2020, pp. 3444--3451, main track.
  [Online]. Available: \url{https://doi.org/10.24963/ijcai.2020/476}
\BIBentrySTDinterwordspacing

\bibitem{22}
\BIBentryALTinterwordspacing
Y.~Li and S.~Lyu, ``Exposing deepfake videos by detecting face warping
  artifacts,'' \emph{CoRR}, vol. abs/1811.00656, 2018. [Online]. Available:
  \url{http://arxiv.org/abs/1811.00656}
\BIBentrySTDinterwordspacing

\bibitem{23}
Y.~Nirkin, L.~Wolf, Y.~Keller, and T.~Hassner, ``Deepfake detection based on
  discrepancies between faces and their context,'' \emph{IEEE Transactions on
  Pattern Analysis and Machine Intelligence}, vol.~44, no.~10, pp. 6111--6121,
  2022.

\bibitem{i3d}
\BIBentryALTinterwordspacing
J.~Carreira and A.~Zisserman, ``Quo vadis, action recognition? {A} new model
  and the kinetics dataset,'' \emph{CoRR}, vol. abs/1705.07750, 2017. [Online].
  Available: \url{http://arxiv.org/abs/1705.07750}
\BIBentrySTDinterwordspacing

\bibitem{thispaper}
\BIBentryALTinterwordspacing
N.~Bonettini, E.~D. Cannas, S.~Mandelli, L.~Bondi, P.~Bestagini, and S.~Tubaro,
  ``Video face manipulation detection through ensemble of cnns,'' \emph{CoRR},
  vol. abs/2004.07676, 2020. [Online]. Available:
  \url{https://arxiv.org/abs/2004.07676}
\BIBentrySTDinterwordspacing

\bibitem{xception}
F.~Chollet, ``Xception: Deep learning with depthwise separable convolutions,''
  2017.

\bibitem{31}
K.~Zhang, Z.~Zhang, Z.~Li, and Y.~Qiao, ``Joint face detection and alignment
  using multitask cascaded convolutional networks,'' \emph{IEEE Signal
  Processing Letters}, vol.~23, no.~10, pp. 1499--1503, 2016.

\bibitem{32}
\BIBentryALTinterwordspacing
V.~Bazarevsky, Y.~Kartynnik, A.~Vakunov, K.~Raveendran, and M.~Grundmann,
  ``Blazeface: Sub-millisecond neural face detection on mobile gpus,''
  \emph{CoRR}, vol. abs/1907.05047, 2019. [Online]. Available:
  \url{http://arxiv.org/abs/1907.05047}
\BIBentrySTDinterwordspacing

\bibitem{33}
F.~Wang, M.~Jiang, C.~Qian, S.~Yang, C.~Li, H.~Zhang, X.~Wang, and X.~Tang,
  ``Residual attention network for image classification,'' in \emph{2017 IEEE
  Conference on Computer Vision and Pattern Recognition (CVPR)}, 2017, pp.
  6450--6458.

\bibitem{29}
\BIBentryALTinterwordspacing
J.~Stehouwer, H.~Dang, F.~Liu, X.~Liu, and A.~K. Jain, ``On the detection of
  digital face manipulation,'' \emph{CoRR}, vol. abs/1910.01717, 2019.
  [Online]. Available: \url{http://arxiv.org/abs/1910.01717}
\BIBentrySTDinterwordspacing

\bibitem{34}
J.~Hu, L.~Shen, and G.~Sun, ``Squeeze-and-excitation networks,'' in
  \emph{Proceedings of the IEEE Conference on Computer Vision and Pattern
  Recognition (CVPR)}, June 2018.

\bibitem{3DAdv}
\BIBentryALTinterwordspacing
R.~Hou, C.~Chen, and M.~Shah, ``An end-to-end 3d convolutional neural network
  for action detection and segmentation in videos,'' \emph{CoRR}, vol.
  abs/1712.01111, 2017. [Online]. Available:
  \url{http://arxiv.org/abs/1712.01111}
\BIBentrySTDinterwordspacing

\bibitem{InflationBenefit}
Y.~Liu, G.~Dwivedi, F.~Boussaid, F.~Sanfilippo, M.~Yamada, and M.~Bennamoun,
  ``Inflating 2d convolution weights for efficient generation of 3d medical
  images,'' 2022.

\bibitem{resnets}
\BIBentryALTinterwordspacing
D.~Tran, H.~Wang, L.~Torresani, J.~Ray, Y.~LeCun, and M.~Paluri, ``A closer
  look at spatiotemporal convolutions for action recognition,'' \emph{CoRR},
  vol. abs/1711.11248, 2017. [Online]. Available:
  \url{http://arxiv.org/abs/1711.11248}
\BIBentrySTDinterwordspacing

\bibitem{CutMixPaper}
\BIBentryALTinterwordspacing
S.~Yun, D.~Han, S.~J. Oh, S.~Chun, J.~Choe, and Y.~Yoo, ``Cutmix:
  Regularization strategy to train strong classifiers with localizable
  features,'' \emph{CoRR}, vol. abs/1905.04899, 2019. [Online]. Available:
  \url{http://arxiv.org/abs/1905.04899}
\BIBentrySTDinterwordspacing

\bibitem{comparePaper1}
\BIBentryALTinterwordspacing
J.~Liu, K.~Zhu, W.~Lu, X.~Luo, and X.~Zhao, ``A lightweight 3d convolutional
  neural network for deepfake detection,'' \emph{International Journal of
  Intelligent Systems}, vol.~36, no.~9, pp. 4990--5004, 2021. [Online].
  Available: \url{https://onlinelibrary.wiley.com/doi/abs/10.1002/int.22499}
\BIBentrySTDinterwordspacing

\bibitem{36}
\BIBentryALTinterwordspacing
A.~V. Buslaev, A.~Parinov, E.~Khvedchenya, V.~I. Iglovikov, and A.~A. Kalinin,
  ``Albumentations: fast and flexible image augmentations,'' \emph{CoRR}, vol.
  abs/1809.06839, 2018. [Online]. Available:
  \url{http://arxiv.org/abs/1809.06839}
\BIBentrySTDinterwordspacing

\bibitem{38}
D.~Kingma and J.~Ba, ``Adam: A method for stochastic optimization,''
  \emph{International Conference on Learning Representations}, 12 2014.

\bibitem{logloss}
L.~Ciampiconi, A.~Elwood, M.~Leonardi, A.~Mohamed, and A.~Rozza, ``A survey and
  taxonomy of loss functions in machine learning,'' 2023.

\bibitem{auc}
Z.~Yang, Q.~Xu, S.~Bao, X.~Cao, and Q.~Huang, ``Learning with multiclass auc:
  Theory and algorithms,'' 2021.

\end{thebibliography}

\end{document}